\documentclass[lettersize,journal]{IEEEtran}
\usepackage{amsmath,amsfonts}
\usepackage{algorithmic}
\usepackage{algorithm}
\usepackage{array}
\usepackage[caption=false,font=normalsize,labelfont=sf,textfont=sf]{subfig}
\usepackage{textcomp}
\usepackage{stfloats}
\usepackage{url}
\usepackage{verbatim}
\usepackage{graphicx}
\usepackage{cite}
\usepackage[hidelinks]{hyperref}
\usepackage{todonotes}
\usepackage{xfrac}
\usepackage{cleveref}
\usepackage{svg}
\usepackage{multirow}
\usepackage{booktabs}

\usepackage{xcolor}

\begin{document}

\title{Why Can’t I See My Clusters? A Precision-Recall Approach to Dimensionality Reduction Validation}

\author{Diede P.M. van der Hoorn, Alessio Arleo, Fernando V. Paulovich
\thanks{All authors are with the Department of Mathematics and Computer Science, Eindhoven University of Technology, The Netherlands.}
\thanks{Manuscript received XX XX, 2025; revised XX XX, 2025.}}

\markboth{Journal of \LaTeX\ Class Files,~Vol.~14, No.~8, August~2021}%
{Shell \MakeLowercase{\textit{et al.}}: A Sample Article Using IEEEtran.cls for IEEE Journals}


\maketitle

\begin{abstract}
Dimensionality Reduction (DR) is widely used for visualizing high-dimensional data, often with the goal of revealing expected cluster structure. However, such a structure may not always appear in the projections. Existing DR quality metrics assess projection reliability (to some extent) or cluster structure quality, but do not explain why expected structures are missing. Visual Analytics solutions can help, but are often time-consuming due to the large hyperparameter space. This paper addresses this problem by leveraging a recent framework that divides the DR process into two phases: a \textit{relationship phase}, where similarity relationships are modeled, and a \textit{mapping phase}, where the data is projected accordingly. We introduce two supervised metrics, \textit{precision} and \textit{recall}, to evaluate the relationship phase. These metrics quantify how well the modeled relationships align with an expected cluster structure based on some set of labels representing this structure. We illustrate their application using t-SNE and UMAP, and validate the approach through various usage scenarios. Our approach can guide hyperparameter tuning, uncover projection artifacts, and determine if the expected structure is captured in the relationships, making the DR process faster and more reliable.
\end{abstract}


\section{Introduction}
\IEEEPARstart{D}{imensionality Reduction} (DR) is a popular exploratory data analysis method to help visualize and interpret high-dimensional datasets by projecting data instances to points in a lower-dimensional space (typically 2D or 3D), retaining, as much as possible, the high-dimensional similarity relationships~\cite{espadoto2019toward}. In the fields of visualization and Visual Analytics, it has been frequently used to support analytical tasks~\cite{nonato2018multidimensional}. However, its popularity has expanded to other fields, where exploratory tasks based on similarity relationships are of great relevance, such as single-cell analysis~\cite{10.1371/journal.pcbi.1012403, Kobak2018TheAO}, protein folding~\cite{Viegas2024}, or atomic structure investigation~\cite{PhysRevX.11.041026}.

Clustering analysis is a common task performed using DR projections~\cite{nonato2018multidimensional, xia2021revisiting}. In particular, in some cases, there is the expectation that the projection will reveal clusters that match known information, such as class labels or prior clustering results~\cite{10.1145/2669557.2669559}. An example of such an expectation would be that a projection of the MNIST dataset~\cite{lecun1999object} would show ten well-separated clusters, one for each digit. However, this expectation is not always met, such as when the projection fails to show clusters or the class labels do not align with the cluster structure in the data~\cite{jeon2023classes, jeonMeasuringValidityClustering2025}. As the DR projections are approximations of how the data are structured in the high-dimensional space and thus may not reflect the underlying structure reliably~\cite{paulovich2024dimensionality}, it is beneficial to understand whether the expected structures are absent because they are not in the data or if it is caused by (parameterization of) the projection.

Current approaches to analyzing DR projections fall short in answering the question of whether an expected structure will be present and why (not), and thus, they cannot help a user understand whether this clustering expectation is valid. The usual DR metrics, such as stress, trustworthiness/continuity, and neighborhood preservation, among others, are not supervised; that is, labels are not used, so they are limited in answering why clusters are not present. In this case, supervised metrics are preferable, such as silhouette~\cite{rousseeuw1987silhouettes} or adaptations of existing unsupervised metrics, like label-trustworthiness and label-continuity~\cite{jeon2023classes}. Supervised methods are still limited in revealing whether the data has no inherent cluster structure from a DR perspective, or if the technique and its parameterization are the problem when expected clusters are not represented. Additionally, these metrics can be hard to interpret. Visual Analytics solutions represent a great improvement in helping analysts investigate such problems~\cite{chatzimparmpas2020t, kwon2017clustervision, cavallo2018clustrophile, cutura2018viscoder}. However, the diversity of DR techniques and large hyperparameter space make the process of checking multiple combinations tedious and usually time-consuming. 

In this paper, we introduce two novel metrics inspired by the well-known \textit{precision} and \textit{recall} concepts of information retrieval~\cite{manning2009intro}, focusing on assessing the presence of cluster structures. Both metrics aim to quantify to the extent to which a DR technique captures the expected cluster structure, giving support to answer the question \textit{``Why can't I see my clusters?''}, or more formally \textit{``Why is my projection not representing the  expected clusters?''}. Their magnitudes are meaningful, and these metrics offer novel insights into DR techniques. 

Our approach takes advantage of the framework introduced by Paulovich et al.~\cite{paulovich2024dimensionality} that splits the DR process into \textit{relationship} phase, where the similarity relationships are modeled, and \textit{mapping} phase, where the data instances are projected into the visual space, with the aim of preserving the similarity relationships. Based on this framework, the proposed \textit{precision} and \textit{recall} metrics can quantify the quality of the modeled relationships with respect to class labels (representing the expected clusters). The key idea is that if the relationships do not capture the expected structure, any subsequent mapping that relies on these relationships will also fail to capture it.
Therefore, a metric that evaluates how faithful the modeled relationships are to the existing labels allows for more insights into whether it is the technique, the parameterization, or the data that causes the unexpected results. 
Aside from providing novel insights, our approach enables the optimization of relationship parameters, allowing the modeled relationships to reflect the expected structure as closely as the data permits.
We validate our approach through four usage scenarios: showing the impact of perplexity on t-SNE projections, estimating the number of neighbors for UMAP, detecting projection artifacts, and identifying mismatches between labels and cluster structure. An implementation of the metrics and documentation is available online~\cite{metricscode}.

\section{Related Work}\label{sec:RelW}
\noindent In more formal terms, Dimensionality Reduction (DR) is the process of mapping each instance $x_i$ of a high-dimensional dataset $X = \{x_1,...,x_N\} \in \mathbb{R}^{m}$ into a point $y_i$ on a lower dimensional visual space, typically 2D or 3D, resulting in $Y = \{y_1,...,y_N\} \in \mathbb{R}^{n=\{2,3\}}$. This process seeks to represent relationships found between data instances of the original high-dimensional space in the visual space~\cite{nonato2018multidimensional, espadoto2019toward}.

Many different DR techniques exist. A common taxonomy~\cite{nonato2018multidimensional, espadoto2019toward} classifies such techniques based on the nature of the relationships they aim to preserve. Global techniques, such as classical MDS~\cite{torgerson1952multidimensional}, ForceScheme~\cite{Tejada_Minghim_Nonato_2003, Minghim_Paulovich_DeAndradeLopes_2006}, or Sammon's Mapping~\cite{sammon1969nonlinear}, aim to preserve overall pairwise distances. Local techniques such as Stochastic Neighbor Embedding (SNE)~\cite{hinton2002stochastic}, t-Student Stochastic Neighborhood Embedding (t-SNE)~\cite{van2008visualizing}, Uniform Manifold Approximation and Projection (UMAP)~\cite{mcinnes2018umap}, ISOMAP~\cite{tenenbaum2000global}, Least-Square Projection (LSP)~\cite{paulovich2008least}, or Locally Linear Embedding (LLE)~\cite{roweis2000nonlinear}, focus on preserving local neighborhoods. Focusing on preserving both types of relationships, Neighborhood Retrieval Visualizer (NeRV)~\cite{venna2010information} defines precision and recall in terms of neighborhood preservation between points in the low-dimensional space and instances in the high-dimensional space, allowing for a decision on the trade-off between local and global preservation. Our approach is also inspired by the precision and recall concepts from information retrieval; however, while in NeRV, such metrics are used as proxies to balance local and global preservation, our formulation focuses on measuring how well the expected data structures are represented.

\vspace{0.1cm}\noindent\textbf{Validation.} When transforming high-dimensional data to a visual layout, information loss occurs, as typically $n \ll m$, and the conveyed relationships are approximations of the relationships present in the high-dimensional space~\cite{nonato2018multidimensional, paulovich2024dimensionality}. Therefore, it is important to validate the quality of such approximations so that misleading artifacts can be detected. The typical approach is based on quality metrics. In their survey paper, Nonato and Aupetit~\cite{nonato2018multidimensional} classify DR quality metrics in global and local measures. An example of a global metric is stress~\cite{kruskal1964multidimensional}. Well-known local metrics are trustworthiness~\cite{kaski2003trustworthiness} and continuity~\cite{kaski2003trustworthiness}. In addition to global and local metrics, Espadoto et al.~\cite{espadoto2019toward} define a third category, known as visual separation metrics or cluster-level metrics~\cite{jeon2023zadu}, which measures how well clusters in high-dimensional data are represented in the visual space. Examples of these metrics are Distance Consistency~\cite{sips2009selecting}, steadiness \& cohesiveness~\cite{jeon2021measuring}, and silhouette~\cite{rousseeuw1987silhouettes} (refer to~\cite{aupetit2016sepme} for a comprehensive study of cluster-level metrics). Another possibility is to classify these metrics based on their level of supervision. Unsupervised metrics such as stress, trustworthiness, and continuity do not consider any labels in their computations. Examples of supervised metrics are silhouette coefficient, neighborhood hit~\cite{paulovich2008least}, label-trustworthiness~\cite{jeon2023classes}, and label-continuity~\cite{jeon2023classes}. 

In a recent paper, Paulovich et al.~\cite{paulovich2024dimensionality} introduced the idea that the DR process can be split into two main phases: \textit{relationship} and \textit{mapping}. The \textit{relationship} phase is where the relationships to be preserved by a DR technique are modeled, for instance, through pairwise distances. The \textit{mapping} phase involves transforming these relationships into visual representations by mapping the data instances to the visual space. In this context, the disadvantage of the previously discussed quality metrics is that they are limited to capturing the quality of the modeled relationships. They usually focus on the quality of the mapping by measuring how well the modeled relationships are represented in the visual space. This makes it difficult to pinpoint the source of problems when a projection does not display visible patterns or expected structures; \textit{``Is the problem in the modeled relationships (or dataset) or in the mapping?"}. Additionally, the values that these metrics produce are not always intuitive to interpret~\cite{paulovich2024dimensionality}. Although helpful when comparing projections, they are hardly useful when assessing the quality of individual projections due to their arbitrary nature~\cite{davi2023eurova} (what does a stress of $0.2$ mean?). Also, some of these metrics are dependent on and sensitive to parameterization, potentially biasing the validation analysis depending on how such parameters are defined and how they are connected with the parameters used by the DR technique (should the $k$ of trustworthiness and continuity be the same as the $perplexity$ for t-SNE or \textit{number of neighbors} of UMAP?). The metrics we propose focus on the modeled relationships without presenting any interdependency with DR parameterization, and their magnitudes reflect meaningful concepts, allowing users to judge the quality of a single projection to assess the information loss in the process.

\vspace{0.1cm}\noindent\textbf{Parameterization.} Another relevant aspect of the correct use of DR techniques is hyperparameterization~\cite{wattenberg2016use}. Some DR techniques, such as UMAP and t-SNE, have parameters that significantly impact the produced projections~\cite{wattenberg2016use, coenenUMAP}, resulting in visual representations that may convey misleading information if the parameters are incorrectly defined. Several approaches have been proposed to estimate these parameters. For example, for t-SNE, several guidelines exist, some of which are dependent on dataset properties~\cite{xiao2023optimizing, gove2022new, kobak2019art}. However, these ``heuristics'' sometimes contradict each other and are not suitable for all types or sizes of datasets. Others propose ways of automating hyperparameter selection based on some measure, such as the KL-divergence for t-SNE~\cite{belkina2019automated, cao2017automatic, de2019optimizing}, or cross-entropy for UMAP~\cite{liu2024hybrid}, silhouette score~\cite{jouilili2024optimizing}, or a combined accuracy metric~\cite{gove2022new}. While some of these approaches do take dataset properties into account, none of them consider the dichotomy of relationship and mapping phases, as previously discussed. So, it is not possible to determine if the suggested hyperparameters define the modeling that best represents the data relationships or the hyperparameters that would result in the best overall optimization. Yet, it would be beneficial to determine appropriate hyperparameter values by considering only the phases they impact, thereby improving the overall understanding and control of the DR process. Additionally, existing ``heuristics'' require the projections to be created, even if they focus on clustering structure, such as the silhouette score. Our approach explicitly considers the dichotomy of relationships and mapping hyperparameterization, and by focusing on the quality of the relationships, it does not require the projections; that is, the mapping phase does not need to be executed to assess the presence of expected structures.
 
\vspace{0.1cm}\noindent\textbf{Visual Analytics.} As DR is a complex process, Visual Analytics (VA) approaches have been suggested to help users understand or select DR techniques or assess the quality of projections and the impact of techniques' parameterization. Several VA tools exist that focus on helping users become aware of projection errors or distortions~\cite{stahnke2015probing, martins2014visual, heulot2013proxilens, seifert2010stress, lespinats2011checkviz, raval2024hypertrix, jeon2021measuring}. t-viSNE~\cite{chatzimparmpas2020t} goes beyond, also supporting the analysis of t-SNE hyperparameter space and projection quality. Clustrophile 2~\cite{cavallo2018clustrophile} also supports the hyperparameter space analysis with a focus on guided clustering analysis. Similarly, VisCoDeR~\cite{cutura2018viscoder} enables hyperparameter space exploration, and additionally, supports the comparison of different DR techniques, just as Clustervision~\cite{kwon2017clustervision} assists users in selecting DR techniques. Embedding comparison methods~\cite{Heimerl2022, Boggust2022, Manz2025} can also be used to inspect DR layouts, but have a broader application domain. Manz et al.~\cite{Manz2025} define a concept called \textit{confusion} that is similar to the \textit{precision} metric we propose, but rather than measuring the relationships, it captures how mixed clusters are in the visual space. While these tools enable users to understand the quality of projections, compare different embeddings, and conduct hyperparameter exploration to gain insight into various techniques, they do not explicitly consider how relationships are modeled. Incorporating our proposed metrics into these tools could provide additional insights into important aspects of dimensionality reduction that were previously not captured. Additionally, as our metrics can be used to automate hyperparameter selection, this could also help reduce the hyperparameter space that needs to be explored. 

\section{Methodology}

\subsection{Overview}
\noindent As mentioned earlier, Paulovich et al.~\cite{paulovich2024dimensionality} recently introduced a framework that serves as the foundation for our proposed metrics. In this framework, the DR process is split into two phases: \textit{relationships}, where the relationships between data instances are represented, and \textit{mapping}, where instances are mapped to points in the visual space seeking to represent the modeled relationships. For instance, UMAP first constructs a weighted neighborhood graph from the high-dimensional space and then embeds this graph into the visual space during the mapping phase.

In more formal terms, the \textit{relationships} are represented using an undirected graph $G = (V, E, \Omega)$ where $V = \{v_1,...,v_N\}$ is a set of $N$ vertices representing the data instances and $E = \{e_{ij} | v_i, v_j \in V\}$ is a set of edges representing the relationships, so that $e_{ij} = (v_i,v_j,\omega_{ij} \in \Omega)$ represents the relationship between $v_i$ and $v_j$ in which $\omega_{ij}: E \rightarrow \mathbb{R}$ is a scalar indicating the strength or magnitude of such a relationship. Here, we are interested in local techniques, so $E$ represents local neighborhoods, and if $(v_i, v_j, \omega) \in E$, $v_i$ is said to be a neighbor of $v_j$, and vice-versa. Specifically, for t-SNE and UMAP, $\omega$ represents the probability of two data instances being neighbors.

Based on this formulation, the \textit{mapping} phase of local DR methods, such as t-SNE and UMAP, can be viewed as a system of attraction and repulsion forces on which the \textit{mapping} process seeks to place (strongly) connected instances close together while positioning disconnected apart. In this system, $e_{ij} \in E$ induces attractive forces between $v_i$ and $v_j$ when they are placed too far apart, while repulsive forces act between vertices that are close but not strongly connected~\cite{bohm2022attraction}. 

Since the modeled relationships determine which instances should appear close or far apart in the visual space, they can be analyzed to assess whether the expected structure could emerge in a projection. In this paper, the expected structure is primarily defined by class labels, although it may also be derived from other sources, such as clustering results. The goal is to determine whether corresponding groups can emerge in the visual space. We transform such a problem into an information retrieval problem, using precision to measure whether a point will (potentially) be placed only near points that share the same label, and recall to measure whether that point will (potentially) be near all points that share the same label.  This approach enables us to attribute missing or misleading clusters either to limitations of the modeled relationships or to distortions introduced by the mapping phase: structure absent from the relationships can not be recovered, while structure that is present but not visualized indicates mapping artifacts. 

\subsection{Precision and Recall}

\noindent In order to derive our metrics and since we are focusing on cluster structures, the first element we need is a reference representing expected clusters, that is, a function $l: V \rightarrow L$ that assigns a label to each vertex, where $L$ is a list of categorical values and vertices with the same label belong to the same cluster. In this paper, we will refer to this reference as \textit{class labels}, though it may represent any type of labeling, such as previous clustering results. Our metrics then quantify how well this expected structure is captured in the relationships.

To define our metrics, we take as inspiration the usual information retrieval definitions of precision ($\sfrac{TP}{(TP+FP)}$) and recall ($\sfrac{TP}{(TP+FN)}$)~\cite{manning2009intro}. We define True Positives ($TP$), False Positives ($FP$), and False Negatives ($FN$) using the graph $G = (V, E, \Omega)$ defined on the \textit{relationship} phase. 

In our formulation, $TP$, $FP$, and $FN$ are specified per vertex $v_i$ as lists of vertices. The True Positives ($TP$) of $v_i$ are the set of adjacent vertices of the vertex $v_i$ that share its label
\begin{equation}\label{eq:TP}
    TP_i = \{ v_j \in \Gamma(v_i) | l(v_i) = l(v_j)\},
\end{equation}
where $\Gamma(v_i) = \{v_j \in V | e_{ij} \in E\}$ denotes the set of neighbors of $v_i$, or the set of vertices connected to $v_i$. In other words, the TP for a vertex $v_i$ is the set of other vertices that this vertex is correctly connected to, as this reflects the expectation that vertices with the same label should form a cluster. The FP of a vertex $v_i$ is defined as the set of adjacent vertices that do not share the same label
\begin{equation}\label{eq:FP}
    FP_i = \{ v_j \in \Gamma(v_i) | l(v_i) \neq l(v_j)\}.
\end{equation}
In other words, the FP for a vertex $v_i$ is the set of vertices that $v_i$ is incorrectly 
connected to, which may result in them being placed close together in the visual space. Since the goal is to separate groups of instances with different labels, it is the set of vertices connected to $v_i$ with a different label. Finally, the FN is defined as 
\begin{equation}\label{eq:FN_edges}
        FN_i = \{v_j \in V \setminus \{v_i\} | l(v_i) = l(v_j) \wedge v_j \notin \Gamma(v_i)\},
\end{equation}
composing the set of vertices that $v_i$ is expected to be connected to. 
In other words, it is the list of vertices that share the same label as $v_i$, yet are not connected to it. Missing such connections may result in vertices with the same label being placed far apart in the visual space.

\vspace{0.1cm}\noindent\textbf{Precision.} Now that these definitions have been provided, \textit{precision} can be formulated. In its classical definition, precision measures the fraction of the retrieved items that are relevant~\cite{manning2009intro}. In our approach, we want to know if a vertex may be positioned near \textit{only} vertices with the same label. To do so, we use the previously provided definitions of true and false positives (\cref{eq:TP}, \cref{eq:FP}). However, instead of following the original definition of precision and only considering the cardinality of such sets, we also employ the weights $\Omega$ of the edges $E$, which serve as the multiplicative factors scaling the attraction forces used in the mapping phase. This is necessary to avoid low weights having the same importance as larger weights, which can lead to the incorrect assumption that all connected vertices will exert the same attraction force, whereas in practice, the magnitude of the forces can vary significantly. Precision for vertex $v_i$ is then defined as the magnitude of correct edges ($TP$), divided by the magnitude of the of all vertices it is connected to ($TP+FP$)
\begin{equation} \label{eq:precision_i}\resizebox{.9\hsize}{!}{$\displaystyle
    \mathcal{P}(v_i) = \begin{cases}
        \dfrac{\sum_{v_j\in TP_i} \omega_{ij}}
        {\sum_{v_j \in TP_i} \omega_{ij} + \sum_{v_j \in FP_i} \omega_{ij}} & \text{if } |TP_i| + |FP_i| > 0, \\
        1 & \text{otherwise}.
    \end{cases}$}%
\end{equation}

If a vertex has high precision, this means that there are few or no neighboring vertices that have a different label. For low precision, the inverse holds. 

\begin{figure}[tbp]
    \centering
    \includegraphics[trim = {0cm, 0.6cm, 0cm, 0.25cm}, clip, width=0.65\linewidth]{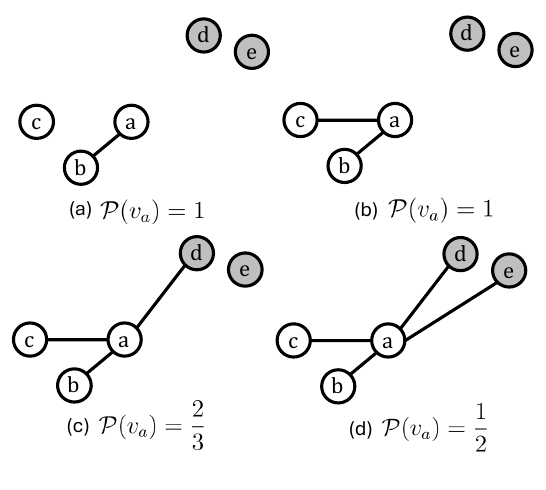}
    \caption{Precision for graphs with different connectivity. Vertices are classified into two classes (white and gray). To keep the example straightforward, only edges that involve vertex $v_a$ are drawn. As the number of edges from $v_a$ to nodes with different labels increases, precision starts to decrease.}
   \label{fig:precision}
\end{figure}

Typically, local DR techniques such as t-SNE and UMAP have a parameter that impacts the neighborhood considered for each data instance, $perplexity$ for the former, \textit{number of neighbors} for the latter; a larger neighborhood typically results in more edges in the graph $G(V, E, \Omega)$. As we increase the number of edges in the graph, a vertex is more likely to become adjacent to a vertex with a different label. This tends to increase the false positives for that vertex, which decreases the precision. This is illustrated in \cref{fig:precision} using a graph with a few vertices and two labels, indicated in gray and white. Also, all edge weights are equal to $1$. As the number of edges in the graph increases, the precision decreases up to the point where $v_a$ is connected to all vertices. 

Considering our formulation, precision tends to be higher if fewer neighbors are considered. So, if only precision is used as a metric to check the quality of the modeled relationships, the result would be biased to small neighborhoods, which has been shown to be problematic for some local techniques~\cite{wattenberg2016use}, resulting in artifacts. Also, the precision of a vertex can only provide information about the ratio of desirable to undesirable edges. It does not consider whether there is enough connectivity between vertices with the same label. Therefore, recall is necessary.

\vspace{0.1cm}\noindent\textbf{Recall.} In its original formulation, \textit{recall} measures what fraction of all relevant instances is retrieved~\cite{manning2009intro}. We transform this into what fraction of vertices with the same label as $v_i$ is connected to the vertex $v_i$. As discussed, to calculate recall ($\sfrac{TP}{(TP+FN)}$), it is necessary to have the True Positives (TP) and False Negatives (FN) of a vertex. These can be computed using \cref{eq:TP} and \cref{eq:FN_edges} previously defined. By using the definition of False Negatives provided in \cref{eq:FN_edges}, and calculating recall as the cardinality of $TP_i$ and $FN_i$ sets, a vertex $v_i$ reaches its maximum recall ($=1$) when $\Gamma(v_i)$ contains all vertices that share its label with $v_i$. However, there might already be sufficient neighborhood connectivity between $v_i$ and all other $v_j$ where $l(v_i) = l(v_j)$ to compose a cohesive cluster without $v_i$ being adjacent to all. Alternatively, we can say that a vertex may be placed in the same cluster as all relevant vertices if it has a connection to them, i.e., if there is a path from the vertex $v_i$ to any other vertex $v_j$ where $l(v_i) = l(v_j)$. 

\begin{figure}[tbp]
    \centering
    \includegraphics[width=0.9\linewidth]{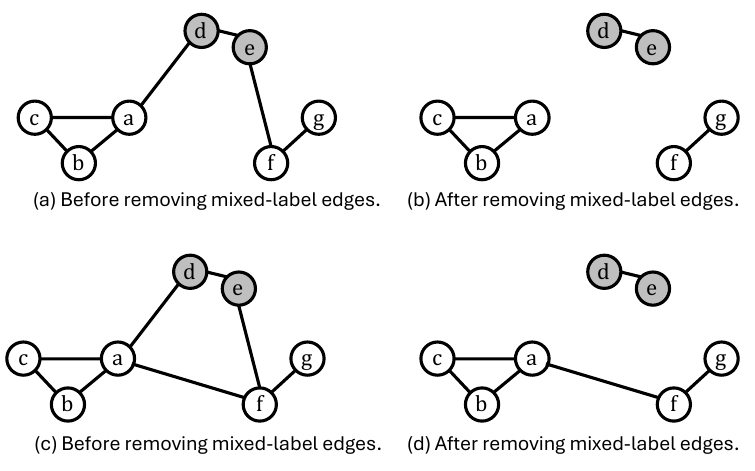}
    \caption{Effect of mixed-label removal on recall in graphs with different connectivity.
    In the top row, removing mixed-label edges creates two separate components for the vertices with the white label, $\{v_a, v_b, v_c\}$ and $\{v_f, v_g\}$. The bottom row shows the same graph with an additional edge $(v_a, v_f)$. Now, after removing mixed-label edges, the result is a single component, $\{v_a, v_b, v_c, v_f, v_g\}.$~This results in higher recall scores ($\mathcal{R}_0(v_a) = 1, \mathcal{R}_1(v_a) =\frac{3}{4}$) than in scenario (a)~($\mathcal{R}_0(v_a) = \mathcal{R}_1(v_a) =\frac{2}{4}$).}
    
   \label{fig:recall}
\end{figure}

For this alternative approach, we define False Negatives based on \textit{connected components} containing vertices with the same label. We create such components by removing any mixed-label edges, the edges between vertices with different labels, so that paths through vertices with different labels are not considered. Thus, we remove any edge $e_{ij} \in E$ that is between two vertices $v_i,v_j$ for which $l(v_i) \neq l(v_j)$ holds.  Notice that by removing the mixed-label edges, every resulting component will only contain vertices sharing the same label, and that multiple components with the same label may be composed. This process of removing mixed-label edges is illustrated in \cref{fig:recall} with two different examples. We use $C(v_i)$ to denote the set of vertices included in the same component of $v_i$, resulting in the following definition of False Negatives 
\begin{equation}\label{eq:FN_component}
        \overline{FN}_i = \{v_j \in V \setminus \{v_i\} | l(v_i) = l(v_j) \wedge v_j \notin C(v_i)\}.
\end{equation}

If we consider the first approach of \cref{eq:FN_edges} as the upper bound of edges required, we can think of this alternative approach as the lower bound, potentially resulting in more or less cohesive clusters. Combining these two approaches of computing False Negatives allows us to characterize recall in terms of the overall connectivity between vertices sharing the same label, varying between the definition in \cref{eq:FN_edges}, which favors high degree vertices, and the definition in \cref{eq:FN_component}, which favors lower degree vertices. This results in
\begin{equation}\label{eq:recall_i}
\resizebox{.9\hsize}{!}{$\displaystyle
    \mathcal{R}_{\alpha}(v_i) = \begin{cases}
        \dfrac{ |TP_i|} {|TP_i| + (\alpha|FN_i| + (1-
        \alpha)|\overline{FN}_i|)} & \text{if } |TP_i| + |FN_i| > 0, \\
        1 & \text{otherwise},
    \end{cases}$}
\end{equation}
where $\alpha \in [0,1]$ is used to balance between the two definitions for the False Negatives (FNs), with larger values of $\alpha$ potentially defining more cohesive representations of clusters in the final projection. Notice that, different from precision, recall is defined based on cardinality since the FNs reflect non-existing edges. Although we could use edge weights to define the TPs, as we did for precision in \cref{eq:precision_i}, for FNs the edge weight is zero (or close to zero), which would result in high recall values even when many neighbors are missing, so we prefer to define recall only based on cardinality.

\begin{figure}[tbp]
    \centering
    \includegraphics[width=0.8\linewidth]{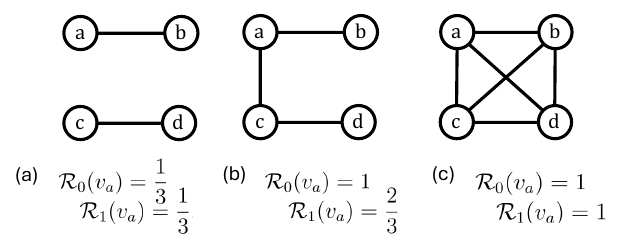}
    \caption{Recall for graphs with different connectivity. There is only a single label, and the group size is $4$. $\mathcal{R}_0(v_a)$ is $1$ as soon as there is a path from $v_a$ to all other vertices, while $\mathcal{R}_1(v_a)$ recall requires an edge to all other vertices to reach $1$. So $\alpha$ can be used to balance lower and higher connectivity degrees.}
   \label{fig:rec_FNs}
\end{figure}

The effect of increasing the number of edges and the resulting recall when $\alpha=0$ and $\alpha=1$ is illustrated in \cref{fig:rec_FNs}. In this example, from (a) to (c), edges are included until there is an edge between all vertices with the same label. Since recall does not consider False Positives, it does not matter whether there are more edges between vertices with different labels. Thus, recall generally has an inverse relationship with precision, and increasing the number of edges tends to increase recall.

\vspace{0.1cm}\noindent\textbf{F-score.} Precision and recall are complementary. In our case, precision measures whether the modeled relationships connect vertices with the same label, whereas recall captures whether vertices with the same label are sufficiently connected. Consequently, a relationship model with high precision tends to have fewer edges (low degree), and a model with high recall tends to be highly connected (high degree). So, relationship models should present a balance between these metrics, resulting in cohesive and pure clusters. In information retrieval, the typical strategy to trade off precision and recall in one single metric is the \textit{f-score}~\cite{manning2009intro}, defined as

\begin{equation}
    f_{\beta}\text{-score}(v_i) =(\beta^2 + 1) \cdot \frac{\mathcal{P}(v_i) \cdot \mathcal{R}_{\alpha}(v_i)}{(\beta^2 \cdot \mathcal{P}(v_i)) + \mathcal{R}_{\alpha}(v_i)},
\end{equation}
where $\beta$ controls the balance between precision ($\mathcal{P}(v_i)$) and recall ($\mathcal{R}_{\alpha}(v_i)$).

While this definition of \textit{f-score} per vertex has some practical use, for analytical purposes, we are most interested in the \textit{f-score} per label and globally. As the former provides information on how well-separated a specific group of vertices sharing the same label is, the latter gives insight into the separation for the whole dataset (all labels). Here, we provide a general definition to determine the global \textit{f-score} as
\begin{equation}\label{eq:fscore}
    f_{\beta}\text{-score}(G) = \frac{1}{|L|} \sum_{l \in L} \left(
    \sum_{v_i \in C_l} \frac{f_{\beta}\text{-score}(v_i)}{|C_l|}\right),
\end{equation}
where $L$ is the set of unique labels, $C_l$ is the set of vertices sharing the same label $l$, that is, ${C_l = \{v_i | l(v_i) = l, v_i \in V \}}$. With this definition, we average the average of the \textit{f-scores} per label, thereby mitigating unbalanced label problems where majority labels dominate the metric. Notice that the \textit{f-score} per label can be computed using this equation by considering only the inner parenthesis summation. 

\begin{figure}
    \centering
    \includegraphics[width=1.0\linewidth]{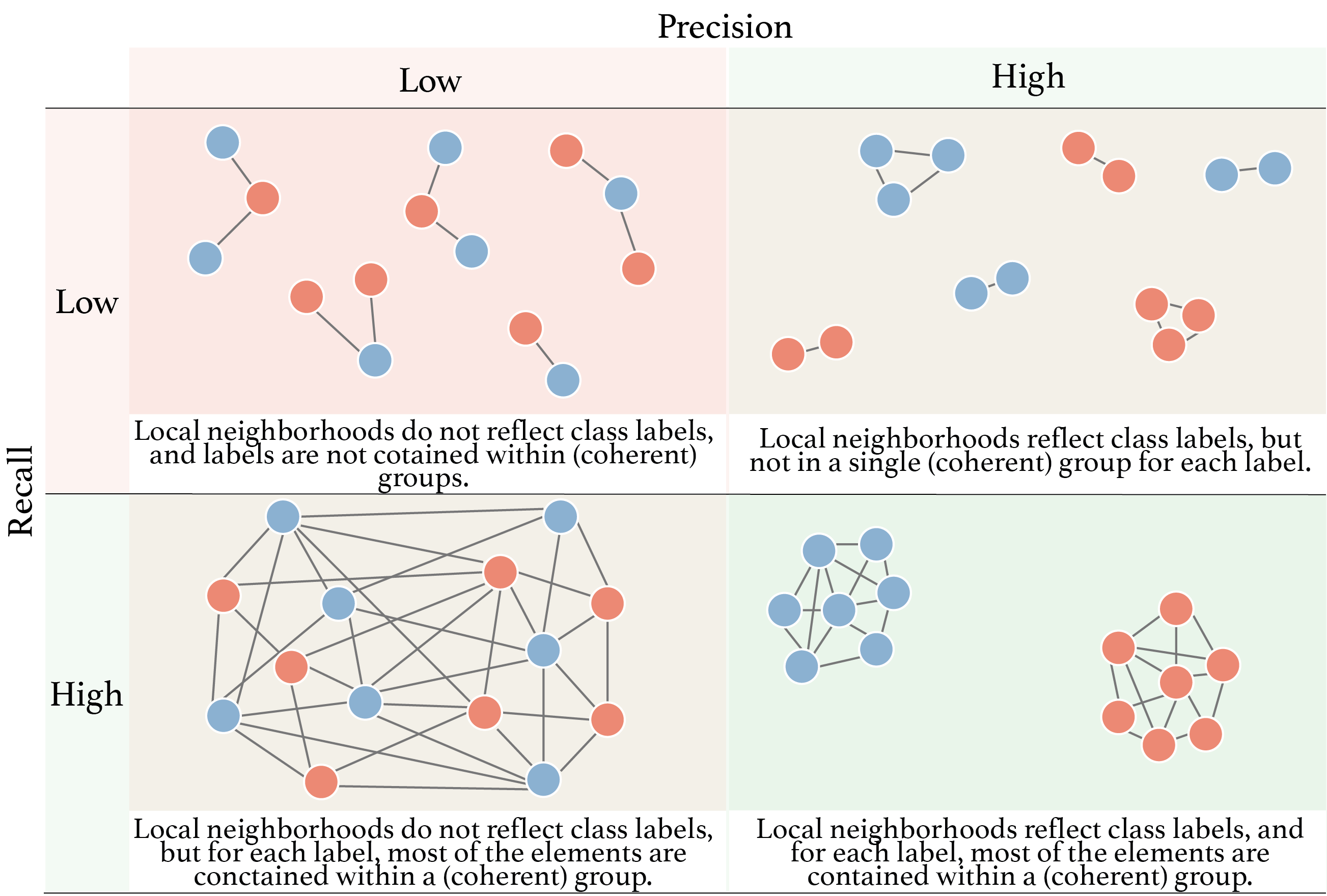}
    \caption{Interpretation of recall and precision. Note that we show the structure of the relationship graph, not a mapping result. Varying $\alpha$ for recall impacts how coherent the groups should be.}
    \label{fig:interpretation}
\end{figure}

\vspace{0.1cm}\noindent\textbf{Interpretation.} \Cref{fig:interpretation} outlines how to interpret different combinations of precision and recall. High f-scores result from both values being high, and the inverse holds for low f-scores. However, as the f-score averages precision and recall, mid-range values can be ambiguous. Therefore, we recommend interpreting precision and recall as separate values, while the f-score is more suitable for estimating DR relationship parameters (see \cref{sec:estimatingDR}). The neighborhood parameter (e.g., perplexity in t-SNE), which controls the neighborhood size, is part of the analytical process and should be considered in the interpretation of these values. For example, low recall is expected for small neighborhood sizes due to the limited number of connections. Since recall generally increases with neighborhood size, low recall at larger neighborhood sizes suggests that instances with the same label are not well connected in the modeled relationships. Conversely, for small neighborhood sizes, a low precision signals an issue, while low precision for a large neighborhood size is expected as mixed-label connections become more likely. \Cref{fig:distill} illustrates this dependence, by showing that both metrics vary with neighborhood size. Additionally, it is important to consider the $\alpha$ parameter, which reflects the expected cohesiveness. 

\subsection{Estimating DR Relationship Parameters} \label{sec:estimatingDR}

\noindent The global $f_{\beta}\text{-score}(G)$ as defined in \cref{eq:fscore} is a metric that measures the degree to which vertices may be placed near only vertices with the same label (precision) while also forming a group with all vertices with the same label (recall). Larger values of \textit{f-score} indicate that in the \textit{mapping} phase, the DR algorithm could represent the clusters defined by the label function $l$ in the projection. This allows us, for instance, to estimate the parameters that affect the relationships (relationship parameters) by looking for values that potentially result in visual clusters, without executing the \textit{mapping} phase. This is most beneficial for parameters that impact the neighborhoods, such as $perplexity$ for t-SNE or \textit{number of neighbors} for UMAP, as they have large ranges of possible values. 

Considering the modeled relationships $G(k)$ as a function of a neighborhood parameter $k$, with larger values indicating larger neighborhoods, the function $f_{\beta}\text{-score}(G(k))$ varying $k$ is, unfortunately, not concave nor convex. Although \textit{recall} of \cref{eq:recall_i} is monotonically increasing, since it grows as $k$ grows, \textit{precision} of \cref{eq:precision_i} tends to decrease, but is not monotonically decreasing for random layouts. Therefore, the maximization of $f_{\beta}\text{-score}(G(k))$ as a function of $k$ is only viable using some metaheuristic~\cite{BOUSSAID201382}, such as simulated annealing or tabu search. In this paper, we employ Bayesian optimization~\cite{garnett_bayesoptbook_2023}, as it has demonstrated effective results for optimizing expensive-to-evaluate black-box cost functions. This approach is often used to estimate hyperparameters of machine learning techniques~\cite{pmlr-v54-klein17a}. Bayesian optimization constructs a posterior distribution of functions that describes the function to be optimized, in this case $f_{\beta}\text{-score}(G(k))$. As the number of observations increases, the algorithm becomes more certain which regions in the parameter space are worth exploring.

\section{Usage Scenarios}
\noindent In this section, we show that our proposed metrics can be a useful analytical tool by providing some usage scenarios and discussing how they help address the gaps identified in the literature. The scenarios provided involve UMAP and t-SNE. For UMAP, we use the internal topology representation to define the graph $G(V,E,\Omega)$. For t-SNE, we employ the edge-removal strategy as described in ~\cite{paulovich2024dimensionality} to create $G$, where edges with probabilities close to zero are removed. Specifically, we sort edges in descending order and retain only those accounting for the first $90\%$ of their total weight. All code used to generate the results can be found in~\cite{metricscode}. Minor differences in the reported results may occur due to the use of stochastic methods and implementation-dependent behavior.

\subsection{Revisiting the impact of perplexity on t-SNE projections}\label{sec:distill}
\noindent Wattenberg et al.~\cite{wattenberg2016use} show that t-SNE's perplexity parameter has a considerable impact on how data is projected onto the visual space. We recreated several of their examples and selected one to illustrate the effectiveness of our metrics. We only include examples with multiple classes and exclude examples with specific topological structures, as this is outside our scope. Please refer to the supplementary materials for all t-SNE projections and UMAP results on the same data. 

\begin{figure*}[t!]
    \centering
    \includegraphics[width=\linewidth]{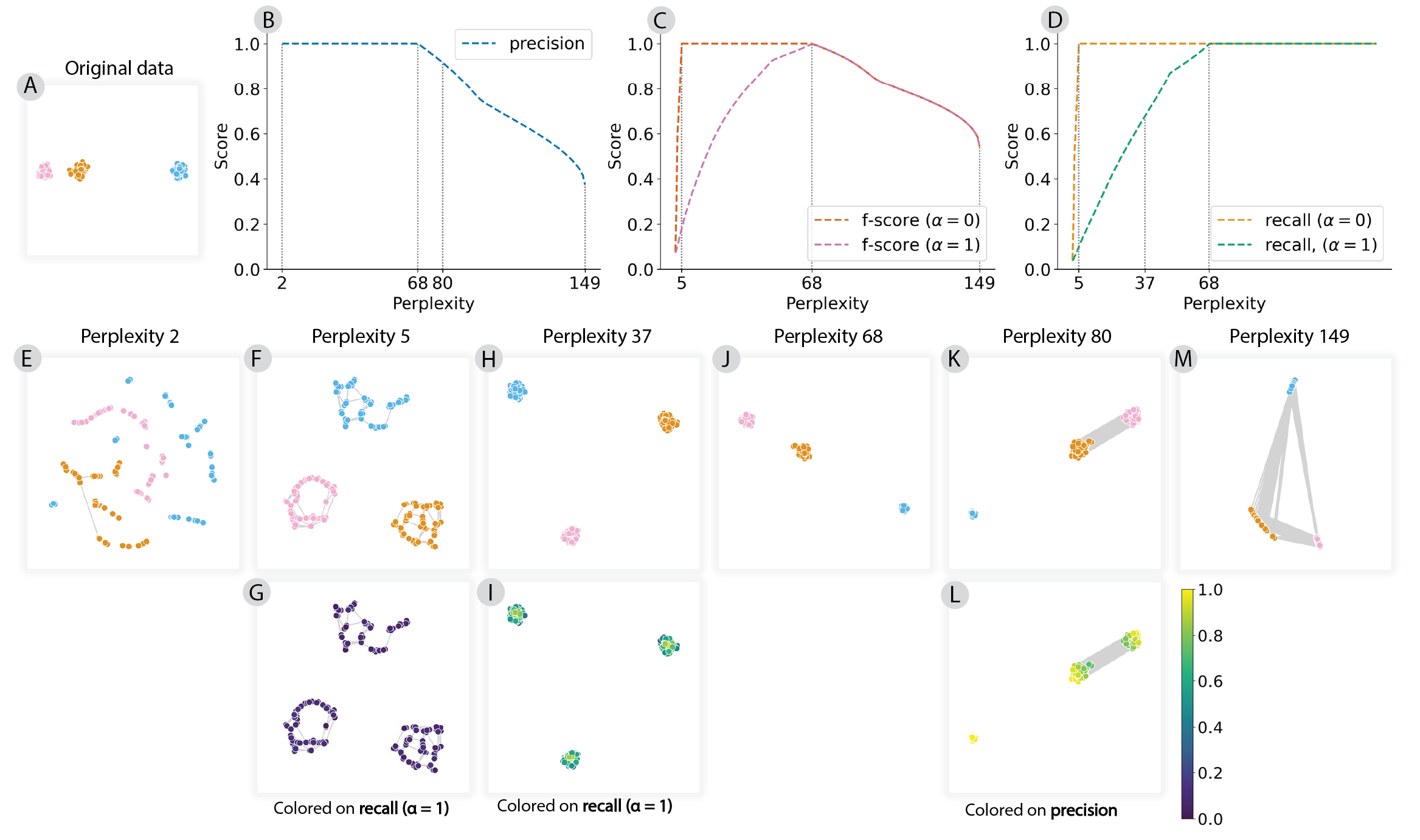}
    \caption{Line charts showing the progression of precision (B), recall (D), and f-score (C) for t-SNE as perplexity increases, along with the resulting projections with their neighborhood graph ((E) - (M), with graph edges represented by gray lines). Original (2D) data is shown in (A). Depending on $\alpha$ (the recall trade-off between higher and lower density clusters), the optimum is reached at perplexity $5$ (F) or perplexity $68$ (J); after this, precision decreases, and edges appear between instances with different labels. (G), (I) and (L) share the same legend. (G) and (I), (L) show the same projection as (F), (H), and (K) respectively, but are colored on metrics instead of class label. For all projections colored on all three metrics please refer to the supplemental material.}
    \label{fig:distill}
\end{figure*}

In this example, we deal with a 2D dataset consisting of $3$ well-separated clusters, each containing $50$ instances (\cref{fig:distill}A). Without requiring the projection to be generated, our metrics can measure the quality of the relationships at any perplexity value. \Cref{fig:distill} shows the progression of precision (B) and recall (D) and the f-scores (C) over a range of perplexities from $2$ to $149$, and use $\beta = 1.0$ in the $f_{\beta}\text{-score}$ (\cref{eq:fscore}) so that we can account for precision and recall equally in this analysis. We present results for $\alpha = 0$ and $\alpha = 1$, ensuring that the full range of possibilities is covered.

We can make several observations based on the line charts shown in \cref{fig:distill}. At $perplexity = 2$, the global precision is $1.0$, as the neighborhood for every vertex is small at this value, resulting only in edges between vertices with the same label. However, while the precision is high, both types of recall are very low, as there are not enough edges between all vertices with the same label. This is reflected in the projection (\cref{fig:distill}E), where each class is projected as many small clusters without edges between them. At $perplexity = 5$, precision remains at $1.0$, but recall with $\alpha = 0$ is also at $1.0$, indicating that each class is contained in a connected component (\cref{fig:distill}F).

As perplexity increases, recall with $\alpha=0$ stays at 1.0 due to larger neighborhoods preserving connectivity (\cref{fig:distill}D). Recall with $\alpha=1.0$ rises until perplexity 68, where it reaches 1.0. At perplexity $37$, roughly halfway between $2$ and $68$, we find that the score for this recall is $0.68$, so on average, the neighborhood of each vertex contains $68\%$ of all vertices with the same label. In the projection colored on recall (\cref{fig:distill}I), we see that the points in the center of the clusters have higher scores, so they are connected to more vertices with the same label than the darker colored vertices, and thus end up in the middle of the clusters. At $perplexity = 68$, all measures are at $1.0$, and if we consider the projection at this point (\cref{fig:distill}J), we can see that the three clusters are well separated.  After $perplexity = 68$, precision starts to decrease, as the edges between vertices with different labels start to increase. Even though for each class the number of points is $50$, it is important to remember that the perplexity is a smooth measure for the number of neighbors and does not reflect the node degree. At $perplexity = 80$, precision is $0.92$, which indicates that $92\%$ of the average edge weights come from vertices with the same label. The remaining come from edges between different labels, in this case orange and pink (\cref{fig:distill}K). \Cref{fig:distill}L shows that points neighboring the edge of the two clusters have lower precision, which can be seen by the fact that these show up as green rather than yellow. These metrics thus do not measure global structure, such as inter-class distances, as this information is not available in the graph. At perplexity $149$, the precision has dropped to $0.37$, indicating that the majority of the edge weights are now between vertices with different labels, as is confirmed by the projection shown in \cref{fig:distill}M. 

To conclude, in line with the paper by Wattenberg~\cite{wattenberg2016use}, we found that perplexity has a considerable impact on the projections. Both too small and too large perplexities can yield unfaithful representations of the original data. Precision and recall are complementary measures that can be used to evaluate this without generating projections. Precision alone may be misleading, as perplexity 2 yields perfect precision but has very poor cluster structure due to the lack of connectivity. Relying only on recall may favor overly large perplexity values (e.g. $149$), without considering that this introduces too many edges. By tuning $\alpha$, it is possible to balance the cluster density. We report the upper and lower bounds to give an idea of the impact on the resulting measures.

\subsection{Estimating Perplexity}

\noindent As mentioned in \cref{sec:distill}, t-SNE is very sensitive to changes in perplexity. Since the range of possibilities for perplexity is quite large, it is common to use the default parameters of existing implementations, such as the popular scikit-learn~\cite{pedregosa2011scikit}. However, projecting the (8-class) \textbf{fiber} dataset~\cite{poco2012employing} using this default parameter ($perplexity=30$) shows multiple subclusters per class (see supplemental material for t-SNE projections, \cref{fig:umap}A and B show a similar result). In this example, we show that our metrics can be used to select an appropriate neighborhood size. The fiber dataset contains $19{,}029$ brain streamlines (fibers) represented as 40-dimensional feature vectors and labeled into 8 classes, each class corresponding to a distinct fiber bundle. Therefore, it would be the expectation that each class is represented by a single cluster, yet this is not the case. Then the question arises: Is cluster fragmentation a parameterization (perplexity) problem, or is it a dataset property? 

In such scenarios, our metrics can be used to estimate a perplexity value that is expected to yield cohesive, well-separated clusters using the f-score. Brute-force search is not feasible for large datasets due to the large hyperparameter space and significant running time, so we use Bayesian Optimization (\cref{sec:estimatingDR}). To balance precision and recall, we set $\beta = 1.0$. It is important to note that this approach is not an exact solution and that there may be multiple projections with the same (or marginally different) quality.

\begin{figure}
    \centering
    \includegraphics[width=\columnwidth]{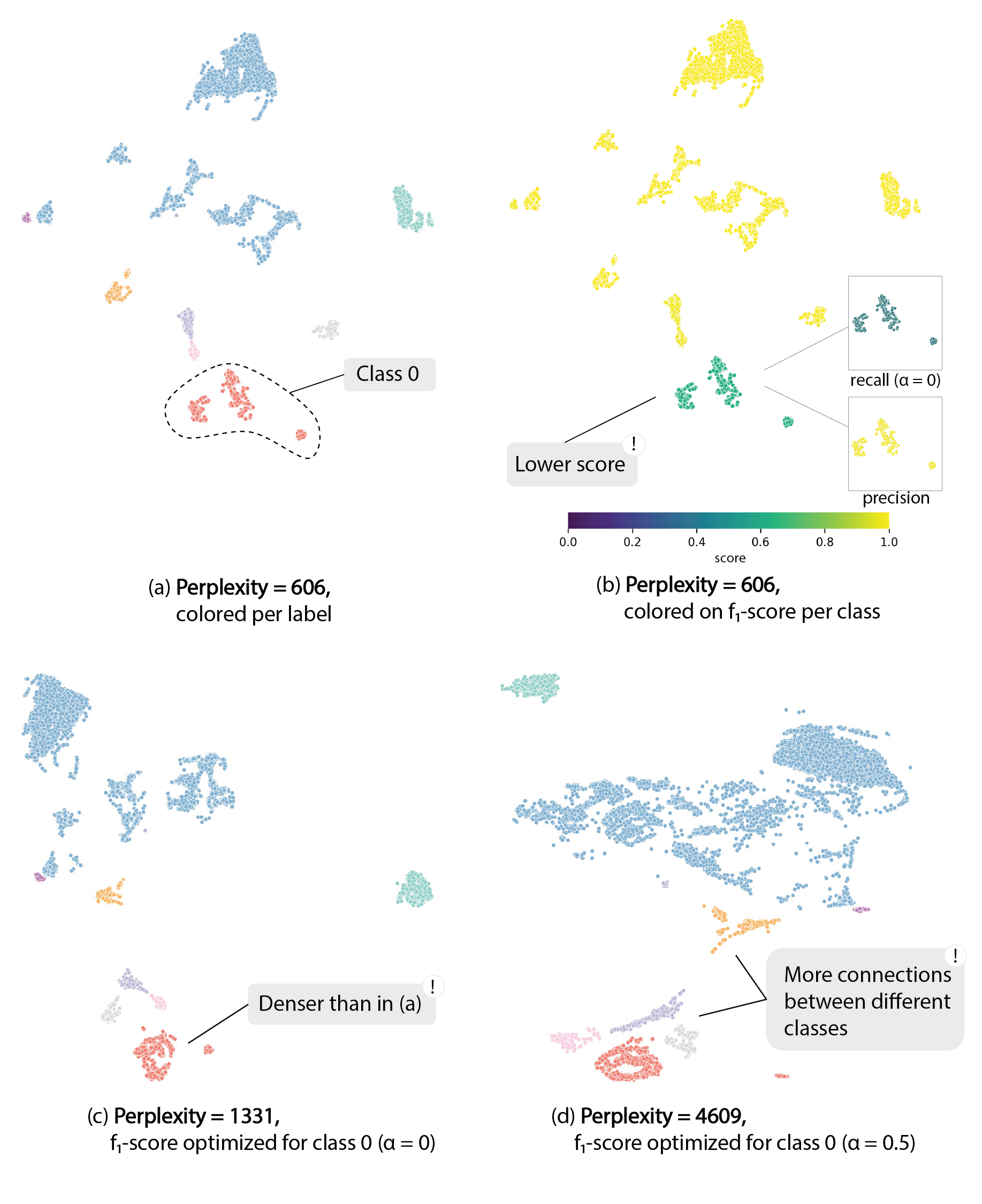}
    \caption{t-SNE projections for the \textbf{fiber} dataset. In (a) and (b), the perplexity is based on the $f_{1}\text{-score}$ for $\mathcal{R}_0$. The low f-score for class 0, caused by low recall, indicates that a different perplexity might be more suitable for this class, as illustrated in (c) and (d). }
    \label{fig:estimation}
\end{figure}

We want to know when each class is contained in a connected component. Therefore, we set $\alpha = 0$ for recall, and we vary perplexity in $[2, 1000]$. The Bayesian Optimization gives a perplexity of $606$ (see \cref{fig:estimation}a)) with an f-score of $0.95$. Since the global f-score averages over each class, it is useful to investigate the score for each class individually. The color coding shows the red class at the bottom has a worse score than the other classes (\cref{fig:estimation}b), which results from a lower recall score. Since $\alpha=0$, it means that not all vertices are contained in a connected component, so we need to increase perplexity to ensure this. By focusing on this class for perplexity estimation, we can learn what perplexity this class needs so that the desired graph connectivity is attained.

As this class likely needs a larger perplexity, we used a range of $[2, 5000]$ to estimate a more suitable perplexity. We considered ${\alpha = \{0; 0.5\}}$ to understand what would happen if we required a denser cluster than a connected component. For $\alpha = 0$, the estimated perplexity is $1,331$ ($f_{1}\text{-score} = 1.0)$, which is much higher than the global estimated perplexity. Keep in mind that this is the $f_{1}\text{-score}$ for the red class; it could be that the global $f_{1}\text{-score}$ is lower if this perplexity introduces too many edges for some classes. For $\alpha = 0.5$, the estimated perplexity is $4,609$ ($f_{1}\text{-score} = 0.87)$. \Cref{fig:estimation}c) and \Cref{fig:estimation}d) show t-SNE projections for these respective perplexities. As we can see, compared to perplexity $606$, the red class becomes denser. However, we also observe that, particularly for the largest perplexity of the two, the gray, pink, and lilac classes no longer form well-separated clusters and instead become much closer to each other, and to the red class. This suggests that the red class is unlikely to ever be fully connected and well-separated, as this would also lead to other classes forming connections and becoming closer. This also explains the lower f-score for the red class; precision decreases as the total strength of relationships to other classes increases. This scenario shows that we need different perplexities to understand different classes properly. For example, we observe that the orange class is starting to spread out again, compared to perplexity $606$. This indicates that the perplexity is too large and that it is connected to other vertices. 

Through our approach, it is possible to find a reasonable estimate of perplexities that best represent each class, as well as a global perplexity. We focus on finding a perplexity based on the relationships, which is the part of the DR process that is impacted by the perplexity. Also, since the f-score is available, this can already be used to gain insight into the potential quality of the embedding. It could be beneficial to combine our approach with other approaches to find parameters for the \textit{mapping} phase, such as optimizing the KL-divergence or silhouette score, but these approaches are not ideal for finding good relationship parameters. Splitting the parameter sets also limits the potential number of combinations that need to be considered. 

\subsection{Finding artifacts}
\begin{figure*}
    \centering
    \includegraphics[width=0.95\linewidth]{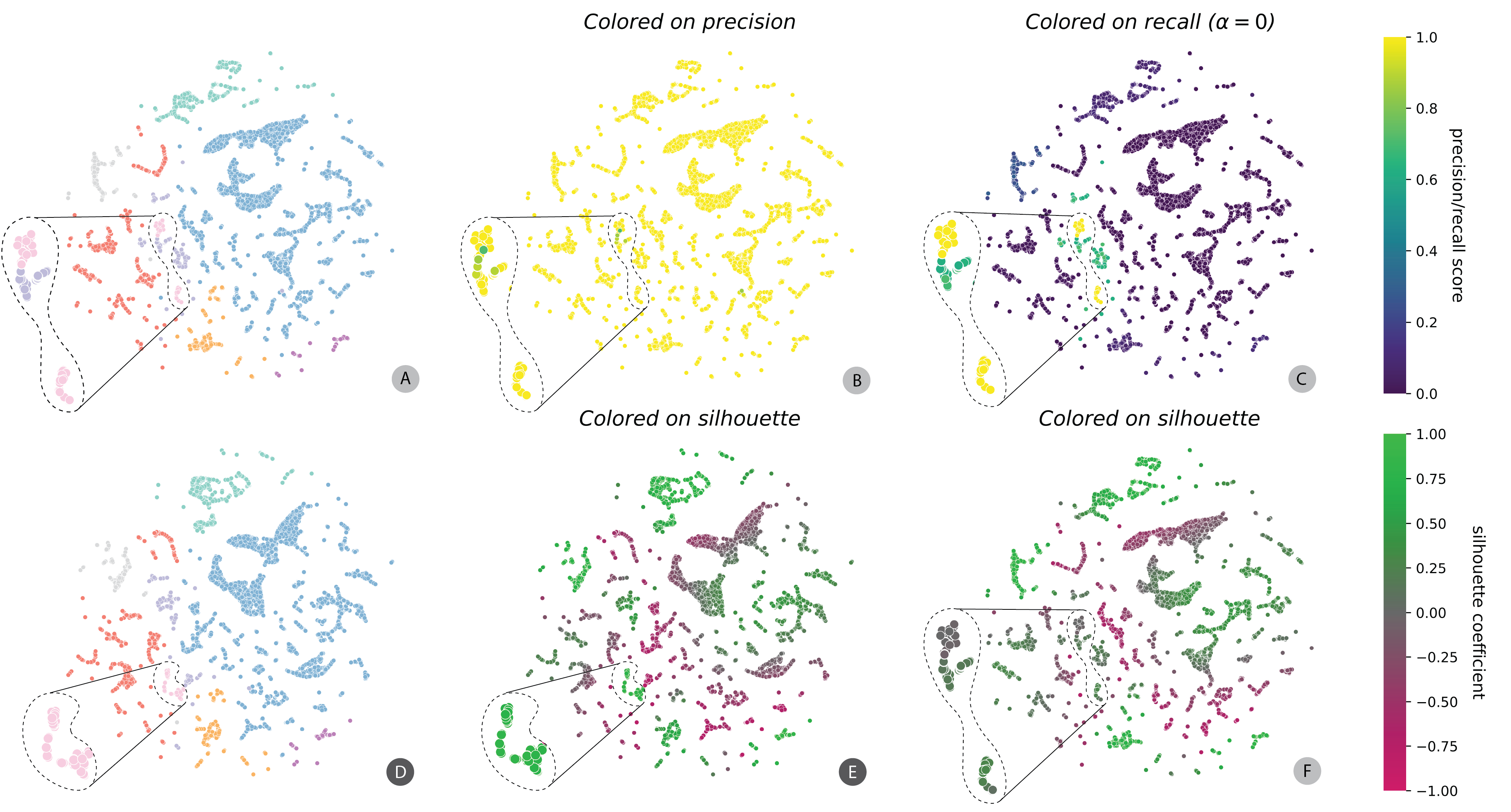}
    \caption{UMAP Projections with visual artifacts,  using the same (default) parameters. Due to the stochastic nature of the algorithm, the outcomes of the mapping phase can vary; (A), (B), (C), and (F) show one version, (D) and (E) another. It appears that the pink class (highlighted in all projections) is split by the lilac class in (A), and while the color coding on precision per point (B) shows that pink and lilac should be close, color coding on recall per point ($\alpha = 0$) (C) indicates that there is a connection between all pink instances, so they should not be split. (D) shows a projection with the same parameters, and indeed, the pink class is now not split by lilac. (E) and (F) show that the silhouette coefficient captures the same issue in terms of clustering quality, but since this is solely based on the projection and not on the relationships, it cannot tell whether it is an artifact, only whether one projection is better than the other.}
    \label{fig:umap}
\end{figure*}
\noindent We evaluate our metrics on the \textbf{fiber} tract dataset. It is well known that the UMAP algorithm is stochastic, and that different runs with the same parameters may yield different results~\cite{coenenUMAP}. In this section, we show that precision and recall enable the identification of artifacts resulting from the \textit{mapping} phase, allowing analysts to diagnose visualization-induced distortions and make more informed judgments about the validity of patterns observed in the embedding.

\Cref{fig:umap} shows the fiber tract dataset, projected with UMAP, using the default parameters ($n\_neighbors = 15$) of the implementation by McInnes et al.~\cite{mcinnes2018umap-software}. Projections are colored by class (A, D) and are evaluated using our metrics: precision (B) and recall with $\alpha = 0$ (C). At first glance, we observe that precision is very high for almost all points, indicating a high proportion of intra-label connections between vertices.

There is a large class composed of multiple sub-clusters, shown in blue in \cref{fig:umap}A and D. This class has a recall score around $0$, as can be inferred from \cref{fig:umap}C. This means that there are not enough edges connecting the items in this class. As we know from the previous example that this class requires quite a large perplexity this is not surprising. Many other classes have a very low recall (\cref{fig:umap}C), meaning they are not in a connected component with many instances of the same class. Together with the high precision, we can conclude that the employed number of neighbors might be too low, as there are hardly any unwanted connections; however, we are missing connections to ensure that every class is at least somewhat connected. The classes with higher recall values typically have a smaller class size, requiring fewer edges to connect the full class.

One class has a very high recall. This class is highlighted in all projections shown in \cref{fig:umap}, and shown as pink in (\cref{fig:umap}A and D). This high recall indicates that all instances in this class are contained within the same connected component. However, if we only look at the projection that is colored using the class label, we might conclude that this class is split into two separate components by the lilac class. Now that we have the information from the recall, it can be seen that the visual representation does not match the metrics. Another UMAP projection with the same parameters (where the relationships also indicate high recall for the pink class) does not show lilac between the pink class (\cref{fig:umap}D). Hence, this is an artifact of the mapping. Now, examining precision, we can see that there are some points with lower precision, meaning they share some similarity with points with a different label, in this case, lilac. Therefore, we also know that the proximity of pink and lilac is not an artifact but is actually captured by the relationships.

\noindent \textbf{Comparing our metrics to Silhouette.} To better situate precision and recall with respect to existing DR quality metrics, we compare them to the silhouette coefficient as this is the most closely related measure. We calculate silhouette on the visual space. While silhouette is designed to evaluate clustering quality, as illustrated in \cref{fig:umap}E and F, it does not measure the difference between the relationship and mapping phases. The pink class has a higher silhouette score in the projection where it is not separated by the lilac class~(\cref{fig:umap}E). However, this score only reflects the quality of the projection and does not indicate whether the projection matches the relationships. Additionally, it is infeasible to generate all possible projections to find the one with the fewest artifacts, and it is not guaranteed that an artifact-free projection exists. With the added information of precision and recall, it is possible to deduce which parts of the projection are reliable.

Applying silhouette directly on the high-dimensional space does not resolve this limitation, as measurements of silhouette on the distances in the original data cannot account for the different perplexities or numbers of neighbors, as distances are not affected by these parameters. Thus, it cannot be used to measure the differences between different model parameterizations. Additionally, silhouette may not be a reliable measure of cluster existence, as the dataset may yield poor silhouette values, while a local DR technique produces a good layout. This is because such techniques transform distances considering neighborhoods.

This raises the question of why we do not apply silhouette on the relationship graph $G(V,E,\Omega)$ directly. The main problem is that, for local techniques, we typically only have distances (or similarities) to a low number of neighbors, whereas silhouette requires distances between all pairs of points. If the distance between disconnected vertices is set to a constant, the distance between disconnected intra-cluster vertices and inter-cluster vertices will have the same value, resulting in poor silhouette values. Consequently, silhouette would only work well for high perplexities where there are connections between most vertices with the same label, introducing a strong bias. Our proposed metrics balance this, considering the trade-off between precision, which tends to low connectivity, and recall, which tends to high connectivity.

\subsection{Classes are not clusters}
\noindent A common DR task is to match clusters with class labels; however, the (implicit) assumption that class labels represent clusters does not always hold~\cite{jeon2023classes, 10.1145/2669557.2669559, jeonMeasuringValidityClustering2025}. Using our metrics, it is possible to verify this assumption. An example where this assumption does not hold is the human activity recognition (\textbf{har}) dataset~\cite{anguita2012human}. This dataset consists of $735$ instances, with $561$ dimensions. There are six class labels, so ideally, there would be six mutually separated clusters in the projection. 

\begin{figure}
    \centering
    \includegraphics[trim = {0cm, 0.35cm, 0cm, 0.4cm}, clip, width= \linewidth]{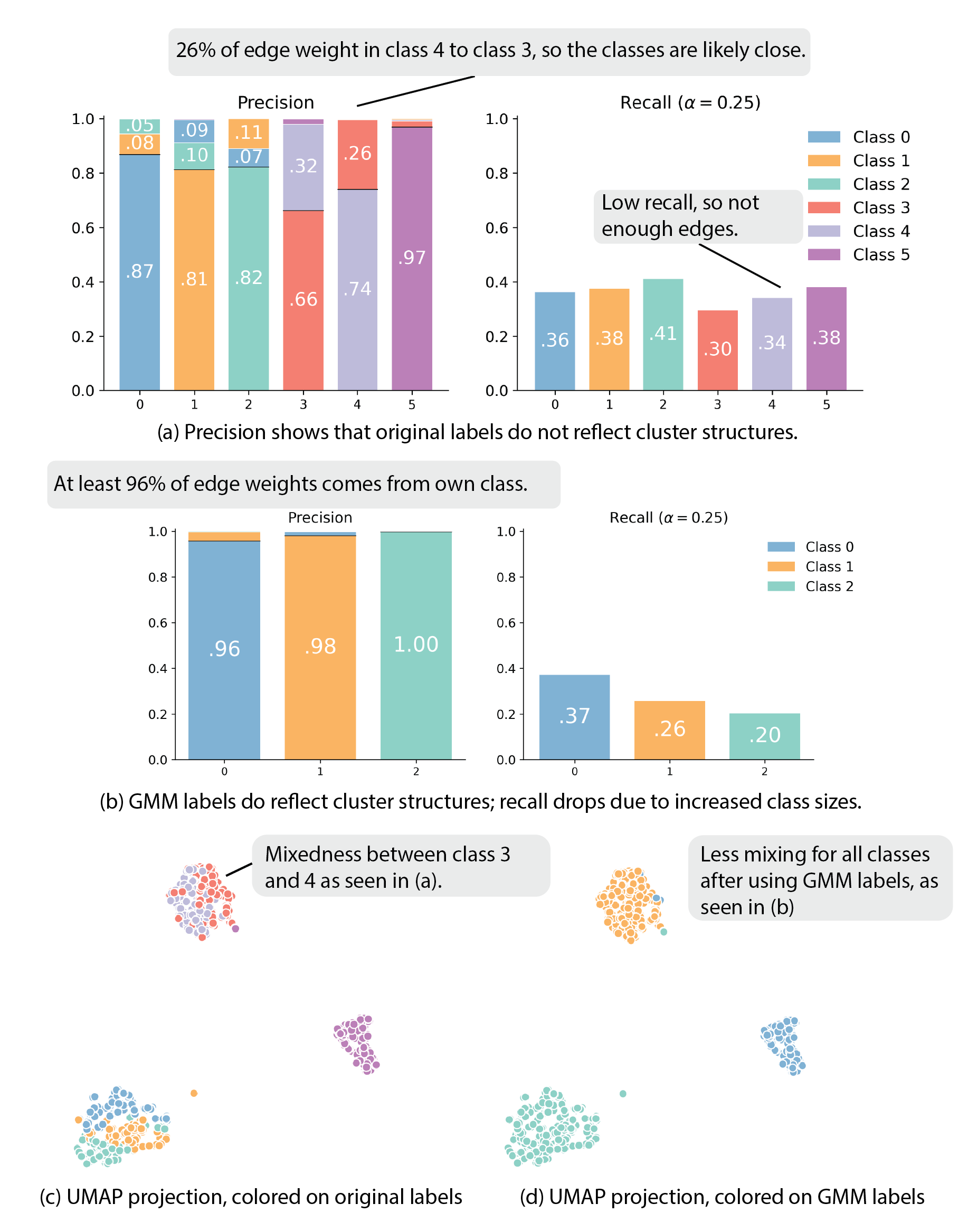}
    \caption{Bar charts showing precision and recall for the \textbf{har} dataset and $n\_neighbors = 15$ (a). The bottom segments of the stacked bar charts for precision represent the true positives; any segment on top of those represents false positives, colored on the class label. Using the clusters created by a Gaussian Mixture Model on the high-dimensional space as class labels is a considerable improvement in terms of precision (b). Recall decreases, as the number of classes decreases, more edges are required. Projections colored on original labels (c) and GMM labels (d) also reflect this.}
    \label{fig:barcharts}
\end{figure}

\Cref{fig:barcharts}a shows bar charts for precision and recall of UMAP, with the number of neighbors set to $15$, as is the default in the UMAP implementation (used here for illustrative purposes). The left chart represents the precision, where each class is represented by a stacked bar. The bottom segment of each bar represents the proportion of (weighted) true positives for that class. The remaining portion represents the false positives, colored per class. A few observations can be made based on this figure. First of all, looking at precision provides the insight that class 5, based on the relationships under this perplexity, is more separated from the others, as 97\% of the edge weights for this class come from its own class. Classes 3 and 4 are also likely to end up close together in the projection, as is the case for classes 0, 1, and 2. Thus, it is possible that there are actually $3$ clusters, rather than the $6$ we expected. We set $\alpha = 0.25$ since we want well-separated clusters, but we do not need them to be very dense. Considering recall, we can see that none of the classes has a very high score, indicating that there are not many connections between the vertices within the same class. Setting $\alpha$ to a lower value would increase the recall score; therefore, if it is acceptable to have fewer connections and more sub-clusters, this can be achieved. From the bar chart results, it is thus possible to deduce that the class labels might not exactly match the clusters. The projection also shows this; while class 5 is relatively well-separated, classes 3 and 4 are mixed, as are classes 0, 1, and 2. The clusters are not very densely clustered, as was expected based on the recall score. 

So now we know that the clusters do not match the labels. Is there any other underlying cluster structure? One approach is to cluster the high-dimensional space. As explained, it appears that there should be $3$ clusters, so we use a Gaussian Mixture Model with $n\_components = 3$ to cluster the data. This approach could also be considered when there are no labels present in the data. The resulting cluster labels can then be used instead labels to compute precision and recall. Using the clustering labels, we observe a considerable increase in precision, as shown in \cref{fig:barcharts}C; however, since the classes are now larger, recall has decreased. If we examine the projection, we can also see that the clustering labels align more closely with the cluster structure. This idea could be further extended by optimizing both the perplexity and the number of clusters in the clustering algorithm. Note that this approach differs from our previous scenario, where we optimized perplexity based on class labels. In contrast, this scenario involves finding labels that best match relationships under a parameterization. In theory, the relationships could also be optimized to reflect the expected cluster structure. However, we do not recommend this approach as this would impose a cluster structure on the data that may not naturally exist. 

Thus, as described by Jeon et al.~\cite{jeon2023classes}, in some cases, class labels might not match the cluster structure defined on the original dataset space given a distance function. Precision and recall can be used to determine this without the need for a projection, and they can be used to find better labels. Note that determining whether this approach is suitable is up to the user. Existing metrics, such as silhouette, are able to determine the clustering quality for one embedding over the other, but since they do not consider the separation between the \textit{relationship} and \textit{mapping} phases, the hyperparameter space may be too large to explore since it is a combination of both \textit{relationship} and \textit{mapping} parameters.

\section{Discussion and Limitations}
\noindent As previously discussed, validation is an essential step for DR, given that the produced projections are approximations of the high-dimensional similarity relationships. However, the typical strategies based on metrics are limited in answering the question \textit{``Why is my projection not representing the expected clusters?''}. The issue is that the existing metrics do not allow for differentiation between whether a failing layout results from a technique limitation or a dataset property. That is the case because such metrics cannot differentiate the impact of the modeled relationships from the mapping phase on the produced projection. Our proposed metrics are novel since they focus specifically on the quality of the modeled relationships. Since the role of the mapping phase is to create visual representations that express the modeled relationships, we can determine whether clusters could potentially be presented based on the data. Through these metrics, it is also possible to determine if the mapping phase is creating artifacts in the final projection, providing support to pinpoint whether bad layouts result from data (modeled relationships) or technique (mapping) limitations. Yet, visual inspection of projections remains a valuable part of the analysis and can reveal insights not captured by these metrics alone. Additionally, they do not consider perceptual grouping, as it cannot be inferred from the relationships, in contrast to perceptually motivated visual quality metrics that analyze the mapping results~\cite{sips2009selecting, aupetit2016sepme}.  

The computational complexity of the metrics is $O(|V| + |E|)$, where the computational complexity of precision is $O(|V|)$, and for recall, it is $O(|V| + |E|)$ due to finding the connected components. Here $|V|$ is the dataset size, and $|E|$ can be at most $|V^2|$, but as the relationship graph is a sparse graph, the time complexity remains closer to linear in $O(|V|)$.
As computational efficiency is a byproduct rather than the goal of these metrics, we have included a running time comparison in the supplemental material. With respect to scalability, both the metrics and the mapping depend solely on the probabilistic graph rather than directly on the original data points. As a result, their behavior depends on the structure on the graph, not explicitly on dataset size, and there is no inherent limitation preventing the application of these metrics to larger datasets. While this suggests good scalability in principle, we leave investigating their behavior for larger datasets for future work.

The proposed precision and recall metrics are easier to interpret and have a clear meaning compared to most of the existing DR metrics~\cite{davi2023eurova}. While precision captures the proportion of correct versus incorrect relationships, recall measures the proportion of missing relationships. The latter also enables capturing within class coherence, beyond class separation (as captured by e.g. silhouette). When $\alpha$ is not set to $0$ or $1$, the quantity expressed by recall becomes more difficult to understand, but still the underlying semantics do not change. The f-score is a different case. Although low values consistently reflect low cohesiveness and separation of clusters, and high values the opposite, the magnitude is not straightforward to interpret. So, our suggestion is to use the f-score for hyperparameter optimization, while precision and recall can be used to investigate the cluster structure. In fact, interpretation is one of the reasons we keep the definitions of precision and recall as simple as possible. We have tried more advanced formulations, particularly for recall, which involve clustering and shortest path graph algorithms. Not only has the meaning of the produced values become harder to interpret, but the running times have also increased substantially. Since no difference in terms of the quality of the measures could be observed, we cannot justify more complex formulations.

There is a challenge in using our metrics that also presents an opportunity: selecting $\alpha$ to balance cluster density and $\beta$ to balance precision and recall in the f-score. Although we discuss meaningful values for both parameters, there remains room for tuning that could lead to improved results. We leave this for future work, since appropriate values must be determined based on the underlying analytical tasks and, potentially, on the technique itself (why did it work better for t-SNE than UMAP?). Addressing these questions requires an extensive study to establish suitable heuristics.

Since we are not performing clustering, DR should reveal only cluster structures actually present in the data, so we do not investigate how to induce a DR method to show clusters, as supported by supervised DR methods~\cite{JMLR:v23:20-188, 10294259, NEURIPS2020_99607461}. Instead, our aim is to assess whether the expected clusters can be reflected in the projection by only considering proper DR technique hyperparameterization, and if not, to understand why. 
Also, the core research question of our paper is not understanding or explaining the results of a projection, as supported by feature-based interpretation strategies~\cite{doi:10.1177/1473871615600010, MARCILIOJR2021115020, TIAN202193, 10.2312:eurova.20231098}, but to give insights about the represented cluster structure or its absence.

For completeness, we also conducted a quantitative analysis to investigate whether there is a correlation between precision and recall and existing DR metrics. Here we summarize the most important findings; for the complete experiments, refer to the supplemental materials. The metrics we compare our metrics to are supervised DR metrics, and since these can behave unexpectedly when datasets do not have well-separated labels~\cite{jeon2023classes}, we do not discuss results for datasets that lack this property. However, for a more comprehensive analysis, we also evaluated how our metrics perform when presented with poorly separated labels. We found a general correlation, particularly for t-SNE, between supervised DR metrics and our metrics. This suggests that our metrics can successfully indicate, without executing the mapping phase, whether a technique, for a specific dataset and parameterization, is able to create a projection that shows the expected cluster separation.
Additionally, we showed that worse label separation results in a decrease in the metric scores. 

\section{Conclusion and Future Work}
\noindent This paper presents novel DR quality metrics inspired by the well-known concepts of precision and recall, aiming to support analysts in providing insight into the reasons why DR layouts fail to display expected cluster structures. The proposed formulations are based on a recent DR framework~\cite{paulovich2024dimensionality}, so they can capture information that is hardly possible to convey through the use of the most popular supervised DR metrics. Our metrics not only provide information to pinpoint whether the absence of expected structures is related to technique or dataset limitations, but also allow the identification of artifacts resulting from the employed approximations, helping to improve DR reliability. In this paper, we examine the proposed metrics for analyzing t-SNE and UMAP. Extending to other local DR techniques and developing heuristics to support the estimation of $\alpha$ and $\beta$, as well as investigating their applicability to larger datasets, are left for future work. In addition, the method could be extended to a semi‑supervised setting by deriving labels through clustering or active learning. In particular, active learning could further enable verification if groupings in the visual space are supported by the modeled relationships.

\bibliographystyle{IEEEtran}
\bibliography{template}

\begin{IEEEbiography}[{\includegraphics[width=1in,height=1.25in,clip,keepaspectratio]{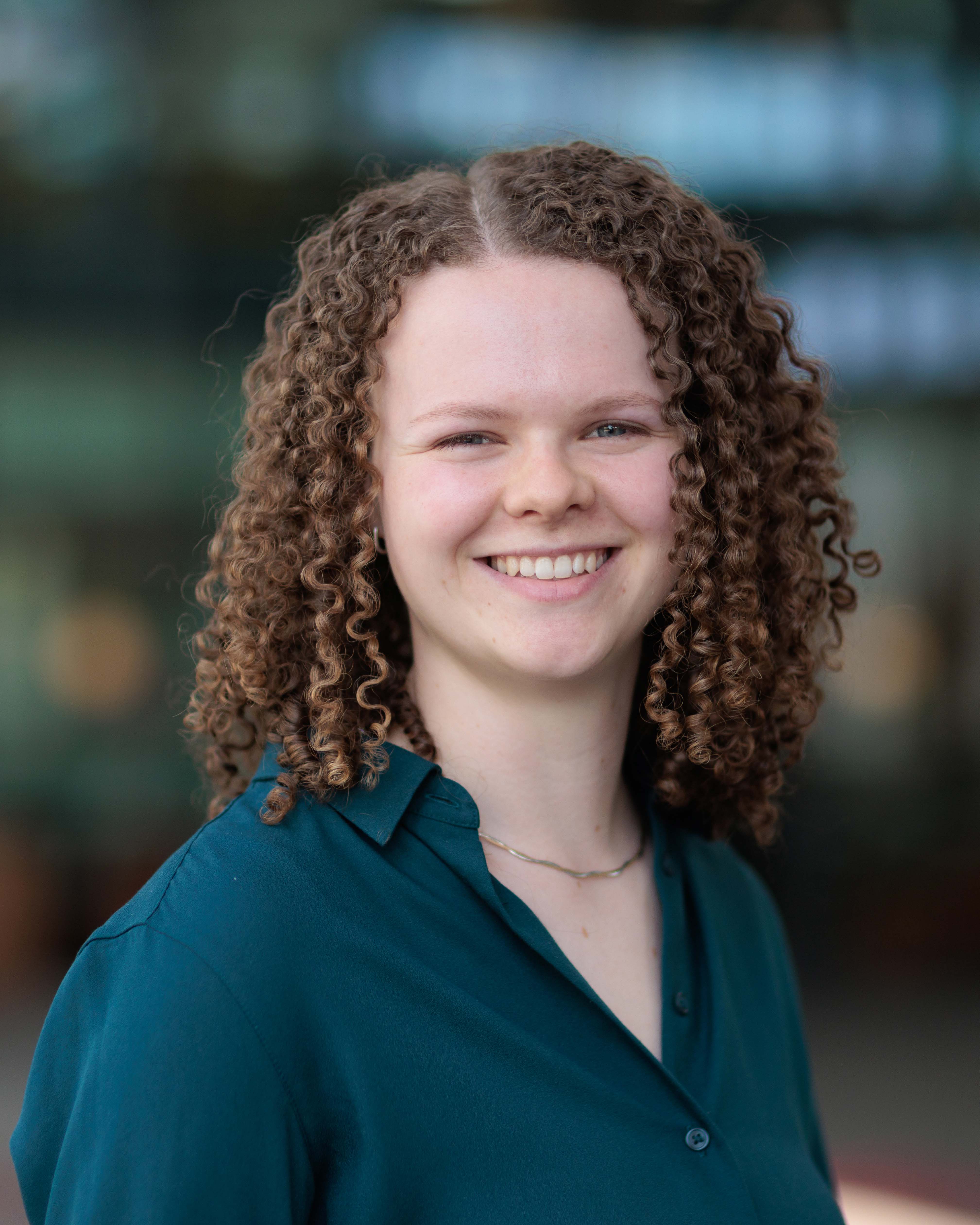}}]{Diede P. M. van der Hoorn}
received her MSc degree in Data Science and Artificial Intelligence from Eindhoven University of Technology. She is currently working toward her PhD degree in the Visualization Cluster within the Department of Mathematics and Computer Science at the Eindhoven University of Technology. Her research interests include dimensionality reduction, the application of graph theory in this domain, and visualization to support the interpretation and analysis of high‑dimensional data. \end{IEEEbiography}

\begin{IEEEbiography}[{\includegraphics[width=1in,height=1.25in,clip,keepaspectratio]{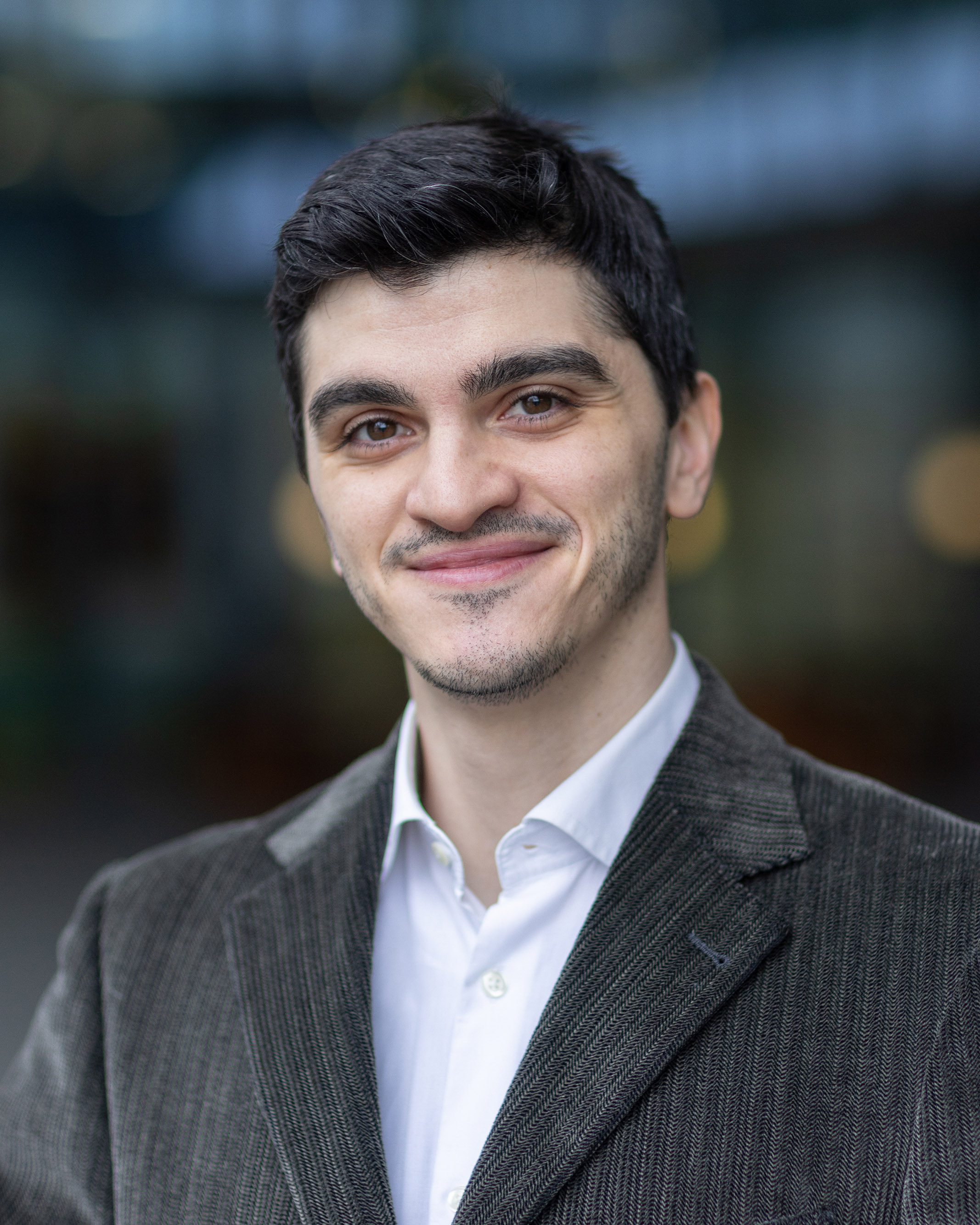}}]{Alessio Arleo} is an Assistant Professor at the Department of Mathematics and Computer Science, Eindhoven University of Technology (TU/e). Previously postdoctoral researcher at Centre for Visual Analytics Science and Technology (CVAST) at TU Vienna. He earned bachelor, master degree (achieved cum laude), and PhD at University of Perugia. His research interests include information visualization, algorithm design, distributed systems, graph drawing, information diffusion, visualization for healthcare.\end{IEEEbiography}

\begin{IEEEbiography}[{\includegraphics[width=1in,height=1.25in,clip,keepaspectratio]{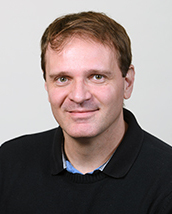}}]{Fernando V. Paulovich} is an Associate Professor in Visual Analytics for Data Science at the Department of Mathematics and Computer Science, Eindhoven University of Technology (TU/e), the Netherlands. Before moving to the Netherlands, he was a Professor and Canada Research Chair at Dalhousie University, Canada (2017-2022), and an associate professor at the University of São Paulo, Brazil (2009-2017). He has been researching information visualization and visual analytics, focusing on integrating machine learning with visualization to leverage machine-learning-enabled automation and user knowledge to make sense of complex, massive datasets.
\end{IEEEbiography}

\end{document}